\documentclass{article}
\usepackage{spconf,amsmath,graphicx}
\usepackage{amssymb,amsfonts}
\usepackage{algorithmic}
\usepackage{textcomp}
\usepackage{xcolor}
\usepackage{booktabs}
\usepackage{multirow}
\def\BibTeX{{\rm B\kern-.00em{\sc i\kern-.025em b}\kern-.08em
    T\kern-.1667em\lower.7ex\hbox{E}\kern-.125emX}}

\usepackage{enumitem}
\setlist{nosep, leftmargin=14pt}

\usepackage{mwe} % to get dummy images

\title{Simulating Post-Neoadjuvant Chemotherapy Breast Cancer MRI \\via Diffusion Model with Prompt Tuning}
\name{Jonghun Kim, Hyunjin Park $^\dag$ \thanks{$^\dag$ Corresponding authors: Hyunjin Park (email: hyunjinp@skku.edu)}}
\address{Department of Electrical and Computer Engineering, Sungkyunkwan University, Suwon, Korea}
\begin{document}
%\ninept
%
\maketitle
\begin{abstract}
Neoadjuvant chemotherapy (NAC) is a common therapy option before the main surgery for breast cancer. Response to NAC is monitored using follow-up dynamic contrast-enhanced magnetic resonance imaging (DCE-MRI). Accurate prediction of NAC response helps with treatment planning. Here, we adopt maximum intensity projection images from DCE-MRI to generate post-treatment images (i.e., 3 or 12 weeks after NAC) from pre-treatment images leveraging the emerging diffusion model. We introduce prompt tuning to account for the known clinical factors affecting response to NAC. Our model performed better than other generative models in image quality metrics. Our model was better at generating images that reflected changes in tumor size according to pCR compared to other models. Ablation study confirmed the design choices of our method. Our study has the potential to help with precision medicine. Our code is available at github.com/jongdory/NAC\_sim.

\end{abstract}


\section{Introduction}
\begin{figure} [t]
    \vspace{-6pt}
    \centering
    \includegraphics[width=0.875\columnwidth]{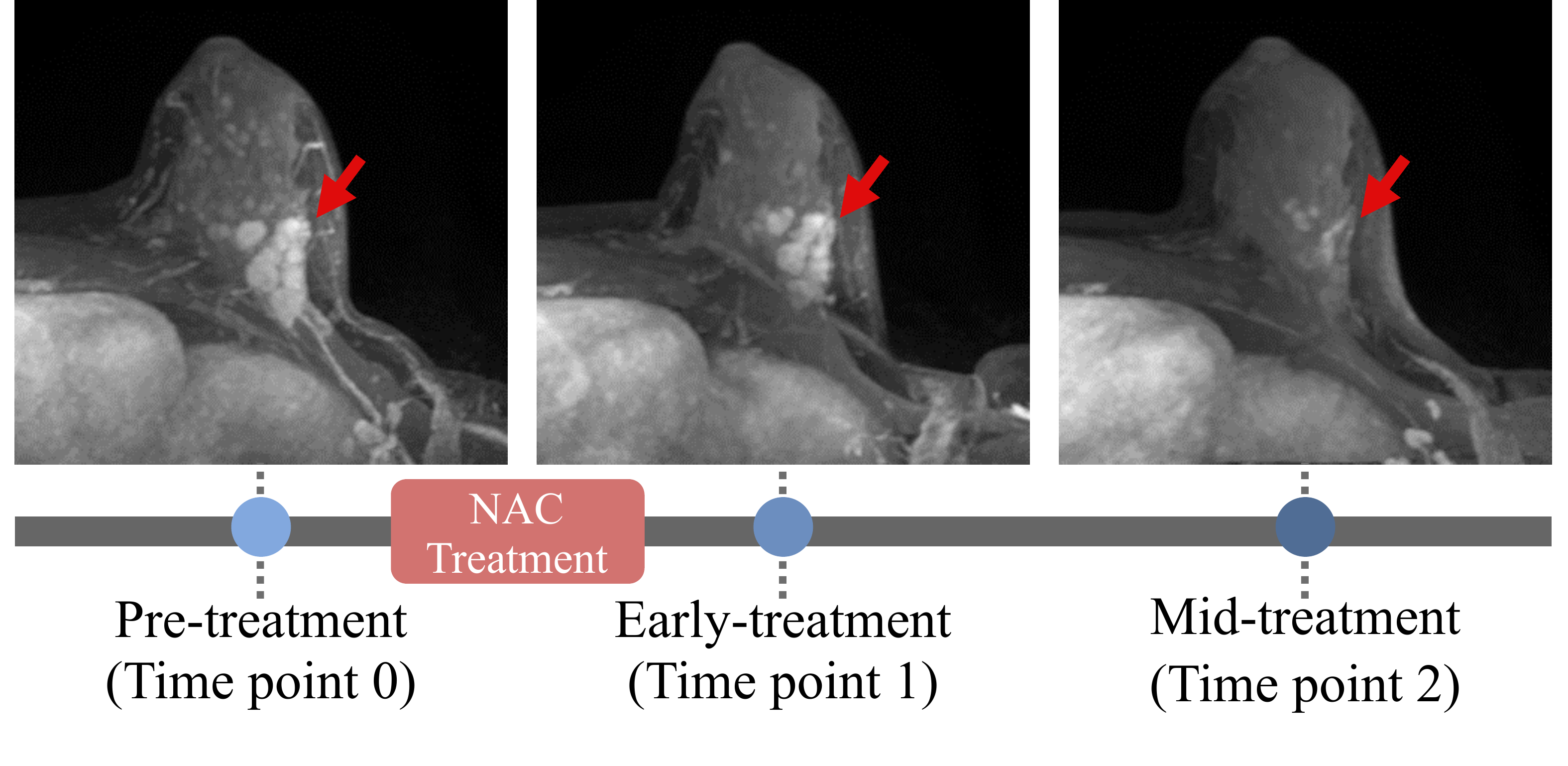}
    \vspace{-18pt}
    \centering
    \caption{Maximum intensity Projection of DCE-MRI with pCR for a patient before and after neoadjuvant chemotherapy treatment. Red arrow points to tumor.} 
    \label{fig1}
    \vspace{-12pt}
\end{figure}

Breast cancer is the second most common cancer in terms of incidence rates worldwide \cite{lei2021global}. Neoadjuvant chemotherapy (NAC) is a common therapy for many patients before the main surgery and does not elicit an immediate response \cite{mieog2007neoadjuvant}. Thus, the response is monitored at follow-up exams at several weeks intervals after NAC with non-invasive imaging. Pathological complete response (pCR) is an established surrogate marker for the long-term outcomes of breast cancer and thus is related to the efficiency of NAC \cite{spring2020pathologic, huang2020association, kim2023multi, 10635671}. Fig. \ref{fig1} depicts images taken before NAC (Time 0), 3 weeks after treatment (Time 1), and 12 weeks after treatment (Time 2). Observing the tumor indicated by the red arrow, pCR patients may show slight tumor shrinkage or no change at the early-treatment stage. By the mid-treatment stage, the tumor size often significantly decreases or disappears entirely. Other factors such as hormone receptor (HR) and human epidermal growth factor receptor 2 (HER2) status are major factors affecting the efficacy of NAC \cite{frenel2023efficacy}. Understanding how breast cancer responds to NAC in hypothesized pCR and non-pCR scenarios can significantly aid in formulating precise treatment plans. 

Dynamic contrast-enhanced magnetic resonance imaging (DCE-MRI) is widely used for non-invasive breast cancer observation by enhancing tumor areas with a contrast agent \cite{yankeelov2007dynamic}. However, its 4D data combining 3D spatial maps and temporal dynamics poses computational challenges for deep learning. To simplify observation, maximum intensity projection (MIP) is often used, as it effectively visualizes tumor regions in contrast-enhanced medical images \cite{kuhl2014abbreviated}. 

% Dynamic contrast-enhanced magnetic resonance imaging (DCE-MRI) has been widely used for the non-invasive observation of breast cancer \cite{yankeelov2007dynamic}. This technique enhances the visibility of breast cancer with the administration of a contrast agent enhancing intensities for tumor-related areas. However, DCE-MRI presents computational challenges for deep learning methods due to the 4D nature of the data combining 3D spatial maps and the temporal dynamics related to contrast agent administration. To simplify the observation of these 3D spatial maps, Maximum Intensity Projection (MIP) is often employed \cite{kuhl2014abbreviated}. MIP is particularly effective in visualizing medical images with contrast agents, as it allows effective observation of tumor-related regions. In this study, we utilize pre-treatment (i.e., before NAC) MIP images from DCE-MRI to simulate follow-up (i.e., after NAC) MIP images in response to NAC using deep learning techniques.

\begin{figure*} [t]
    \vspace{-12pt}
    \centering
    \includegraphics[width=\textwidth]{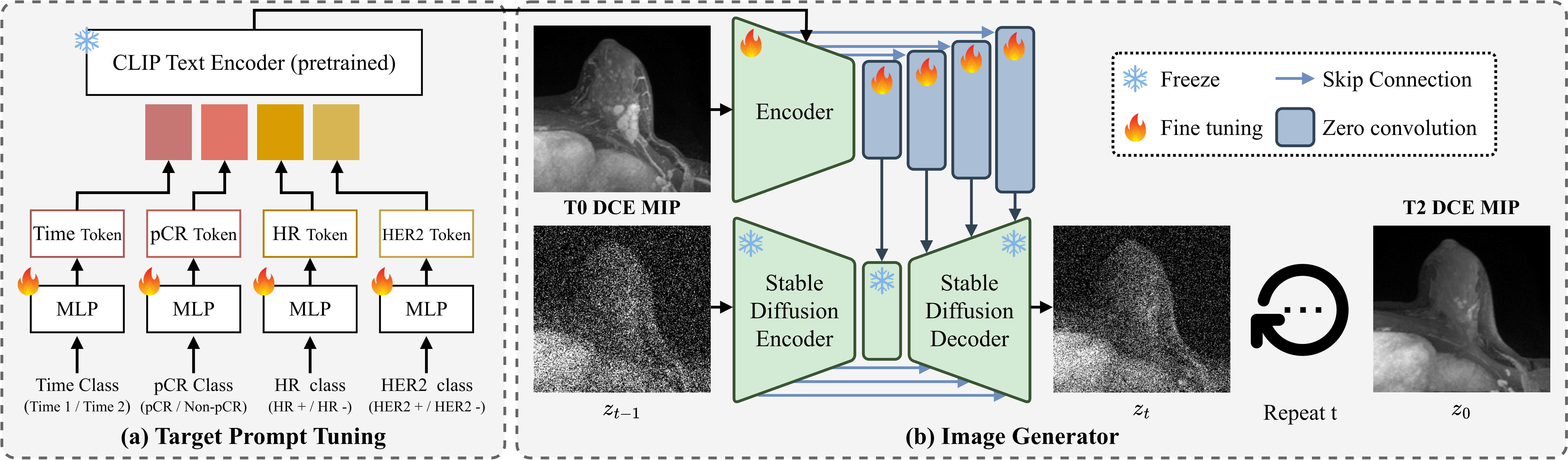}
    \vspace{-21pt}
    \caption{Illustration of a proposed pipeline for simulating images after NAC. (a) Prompt tuning to reflect the patient's clinical information and target time point in the image. (b) Image generator synthesizing the image at target time using pre-treatment images and target prompts through a diffusion model.} 
    \label{fig2}
    % \vspace{-3pt}
\end{figure*}

To predict images over time in response to NAC using deep learning, generative models are suitable. Diffusion models \cite{ho2020denoising} have recently emerged as a promising alternative to generative adversarial networks \cite{goodfellow2014generative}, gaining significant traction \cite{dhariwal2021diffusion}. There is a growing body of research utilizing diffusion models for image synthesis tasks in the medical domain \cite{Kim_2024_WACV, na2024radiomicsfill}. In this study, we simulate MIP images derived from DCE-MRI after NAC using pre-treatment images with diffusion model. To the best of our knowledge, this approach is among the first of its kind in the context of breast imaging. Due to the limited samples in medical imaging (and thus breast imaging), we leverage a pretrained model to train our deep learning model. We employ prompt tuning to account for major factors such as HR, HER2, and pCR during the generation process. Our method can simulate both pCR and non-pCR scenarios effectively. Additionally, to quantify changes in tumor size after NAC, we utilize nnUNet \cite{isensee2021nnu} segmentation method to monitor the changes in tumor regions. 

% To predict images over time in response to NAC using deep learning, generative models are suitable. Generative Adversarial Networks (GANs) \cite{goodfellow2014generative} have long been established as a key model in image synthesis. For tasks like ours, models such as pix2pix \cite{isola2017image}, which facilitate image-to-image translation, have been widely used, and modality translation based on these models has been proposed in medical imaging \cite{dalmaz2022resvit}. However, diffusion models \cite{ho2020denoising} have recently emerged as a promising alternative to GANs, gaining significant traction as generative models \cite{dhariwal2021diffusion}. There is a growing body of research utilizing diffusion models for image synthesis tasks in the medical domain\cite{Kim_2024_WACV, na2024radiomicsfill}.

% In this study, we simulate MIP images derived from DCE-MRI after NAC using pre-treatment images using a diffusion model. To the best of our knowledge, this approach is among the first of its kind in the context of breast imaging. Due to the limited samples in medical imaging (and thus breast imaging), we leverage a pretrained model to train our deep learning model. We employ prompt tuning to account for major factors such as HR, HER2, and pCR during the generation process. Our method can simulate both pCR and non-pCR scenarios effectively. Additionally, to quantify changes in tumor size after NAC, we utilize nnUNet \cite{isensee2021nnu} segmentation method to monitor the changes in tumor regions.
\section{Method}
In this study, we used ControlNet \cite{zhang2023adding}, a diffusion model framework, to simulate post-treatment MIP images after NAC, generating Time 1 and Time 2 images from pre-treatment images (Time 0). Using a pretrained Stable Diffusion model, we fine-tuned the control encoder. Since the CLIP \cite{radford2021learning} text encoder in Stable Diffusion lacks breast imaging expertise, we applied prompt tuning \cite{jia2022visual}, conditioning the text with relevant medical information to improve accuracy and generate precise images.

% In this study, we utilized ControlNet \cite{zhang2023adding}, a diffusion model-based framework, as the backbone for simulating post-treatment MIP images after NAC. Our objective is to synthesize Time 1 images three weeks after treatment and Time 2 images 12 weeks after treatment using pre-treatment images (Time 0). ControlNet generates the target image using a control image and a target prompt. We leveraged a pretrained Stable Diffusion \cite{rombach2022high}, fine-tuning only the control encoder. Stable diffusion employs a pretrained CLIP \cite{radford2021learning} text encoder trained on natural image captions, which lacks the domain knowledge needed for breast imaging. To address this, we implemented prompt tuning \cite{jia2022visual}, conditioning the text prompt with relevant target image information, enabling accurate image generation for the medical domain.

\subsection{Prompt Tuning}
\label{sec:3.1}
We utilize prompt tuning \cite{jia2022visual} to enable conditioning on the target image without performing prompt engineering or fine-tuning the text encoder on our small dataset. Additionally, unlike other models that require separate training for each target time point image (i.e., Time 1 and Time 2 ), our method allows a single model to simultaneously learn and generate multiple target time point images. This method is depicted in Fig \ref{fig2}(a). We create separate class tokens for the time point, pCR, HR, and HER2 of the target image:
\begin{equation}
    \begin{aligned}
    \textit{T}_\text{tp} = \textbf{MLP}_\text{tp}(y_\text{tp}), \textit{T}_\text{pcr} = \textbf{MLP}_\text{pcr}(y_\text{pcr}) \ \ \ \ \\
    \textit{T}_\text{hr} = \textbf{MLP}_\textit{hr}(y_\text{hr}), \textit{T}_\text{her2} = \textbf{MLP}_\text{her2}(y_\text{her2}),  
    \end{aligned}
\end{equation}
where \textbf{MLP} denotes a multi-layer perceptron, $\textit{T}$ represents a token, and $y$ is the class label. The subscripts tp, pcr, hr, and her2 correspond to the time point, pCR, HR, and HER2, respectively.
These tokens are concatenated to form a prompt sequence $\mathcal{P}$, which is fed into the pretrained CLIP text encoder $\text{CLIP}_\text{text}$: 
\begin{equation}
    \mathcal{P} = \text{concat}(\textit{T}_\text{tp}, \textit{T}_\text{pcr}, \textit{T}_\text{hr}, \textit{T}_\text{her2})
\end{equation}
\vspace{-12pt}
\begin{equation}
    c_t = \text{CLIP}_\text{text}(\mathcal{P})
\end{equation}
During training, the MLP is fine-tuned while the $\text{CLIP}_\text{text}$ remains frozen.

\subsection{Training Process}
We aim to generate target time point images given the pre-treatment image, the target time point, and the patient's pCR, HR, and HER2 information. The patient's clinical data and target time point information are fed into the model through prompts generated in section \ref{sec:3.1}, while the pre-treatment image is fed to ControlNet and used as a condition for the denoising diffusion model. The encoding of pre-treatment image $c_{pre}$ is given by:
\begin{equation}
    c_f = \mathcal{E}(c_{pre})
\end{equation}
The image diffusion algorithm progressively adds noise to create a noisy image $z_t$ from an input image $z_0$, where $t$ represents the number of times noise has been added. The model is trained to predict the noise at a given time point $t$ and progressively remove this noise to synthesize the image $z_0$, as illustrated Fig \ref{fig2}(b). To denoise the image, the model takes the condition feature $c_f$ and the target prompt $c_t$ as inputs, using them to remove the noise for synthesizing the target image. The training objective is defined as follows:
\begin{equation}
    \mathcal{L} = \mathbb{E}_{\boldsymbol{z}_0, t, \boldsymbol{c}_t, \boldsymbol{c}_f, \epsilon \sim \mathcal{N}(0,1)}\big[||\epsilon - \epsilon_\theta(\boldsymbol{z}_t, t, \boldsymbol{c}_t, \boldsymbol{c}_f)||^2_2\big],
\end{equation}
where $\mathcal{L}$ is the overall learning objective of the diffusion model. $\boldsymbol{z}_0$ is the target time point image, $\boldsymbol{c}_t$ is target prompt, and $\boldsymbol{c}_f$ is encoded condition image (pre-treatment image). 

\begin{figure*} [t]
    \vspace{-9pt}
    \centering
    \includegraphics[width=0.825\textwidth]{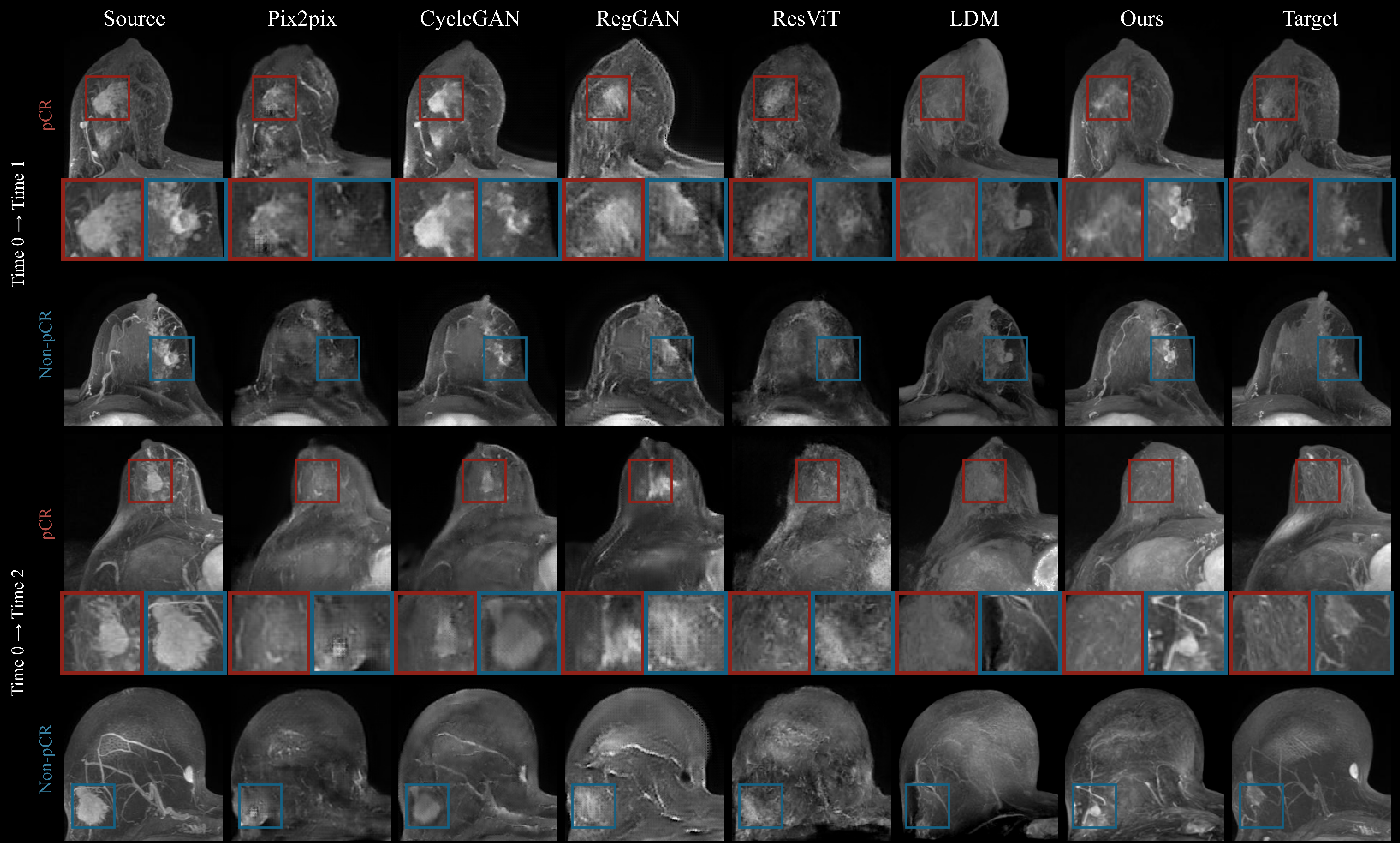}
    \vspace{-12pt}
    \caption{Illustration of generated images after NAC for various methods. Top: Results for the Time 0 $\rightarrow$ Time 1 task. Bottom: Results for the Time 0 $\rightarrow$ Time 2 task. The bounding boxes indicate tumor areas, with enlarged views.} 
    \label{fig3}
\vspace{-12pt}
\end{figure*}

\setlength{\tabcolsep}{3.5pt}
\begin{table} [t]
    \centering
    \caption{Comparison of our model's performance for Time 0 to Time 1 generation task. Bold denotes the best performance.}
    \label{table1}
    \centering
    \vspace{3pt}
    \scalebox{0.65}{
        \begin{tabular}{ccccccccc}
            \toprule
            Task & \multicolumn{8}{c}{Time 0 $\rightarrow$ Time 1}\\
            \cmidrule(lr){2-9} 
            pCR & \multicolumn{4}{c}{pCR} & \multicolumn{4}{c}{Non-pCR} \\
            \cmidrule(lr){2-5} \cmidrule(lr){6-9}
            Model & PSNR $\uparrow$ & SSIM $\uparrow$ & FID $\downarrow$ & LPIPS $\downarrow$ & PSNR $\uparrow$ & SSIM $\uparrow$ & FID $\downarrow$ & LPIPS $\downarrow$ \\
            \midrule
            \multirow{1}{*}{Pix2pix}  \cite{isola2017image}       & 18.50 & 0.528 & 256.0 & 0.216 & 17.94 & 0.449 & 220.8 & 0.269 \\
            \multirow{1}{*}{CycleGAN} \cite{zhu2017unpaired}      & 19.53 & 0.542 & 232.1 & 0.223 & 18.19 & 0.465 & 207.7 & 0.276 \\
            \multirow{1}{*}{RegGAN}   \cite{NEURIPS2021_0f281810} & 18.11 & 0.447 & 233.6 & 0.247 & 17.65 & 0.414 & 205.1 & 0.286 \\
            \multirow{1}{*}{ResViT}   \cite{dalmaz2022resvit}     & 18.92 & 0.507 & 244.9 & 0.222 & 18.07 & 0.432 & 208.6 & 0.278 \\
            \multirow{1}{*}{LDM}      \cite{rombach2022high}      & 17.99 & 0.501 & 226.5 & 0.196 & 17.07 & 0.431 & 195.5 & 0.257 \\
            \multirow{1}{*}{\textbf{Ours}} & \textbf{20.27} & \textbf{0.561} & \textbf{175.4} & \textbf{0.186} & \textbf{18.68} & \textbf{0.467} & \textbf{156.4} & \textbf{0.253} \\
            \bottomrule
        \end{tabular}
    }
    \vspace{-6pt}
\end{table}

\subsection{Inference Process}
To generate the desired target time point image, we first create the target prompt. The target time point can be either Time 1 or Time 2, and clinical information such as pCR, HR, and HER2 status can be set following the patient's clinical report. Using this information, we generate the prompt and use it along with the pre-treatment image in the diffusion model as conditions. We sample $\boldsymbol{z}_T$ from a Gaussian noise distribution $\mathcal{N}(0,1)$, and then use the diffusion model to denoise $\boldsymbol{z}_T$, synthesizing the target time point image (as shown Fig \ref{fig2}(b)).

\section{Experiments}

\noindent \textbf{Implementation Details.}
We used ControlNet v1.1 \cite{zhang2023adding} as the backbone model, which is based on Stable Diffusion 2.1 \cite{rombach2022high}, and utilizes OpenCLIP ViT-H/14 \cite{radford2021learning}. Given the text embedding dimension of 1024 in this model, the MLP for prompt tuning also has a dimension of 1024. Our training was conducted with a batch size of 8 and a learning rate of 5e-6 using the AdamW \cite{loshchilov2017decoupled} optimizer. We set the time steps $T$ to 1000 for training. For inference, we used 200 steps with a DDIM \cite{song2020denoising} to enable faster sampling.

% \subsection{Comparison Models and Evaluation Metrics}
% To validate the effectiveness of our model, we compared it with several commonly used methods in image-to-image translation. We employed Pix2Pix \cite{isola2017image}, CycleGAN \cite{zhu2017unpaired}, RegGAN \cite{NEURIPS2021_0f281810}, ResViT \cite{dalmaz2022resvit} and latent diffusion model (LDM) \cite{rombach2022high} as comparison methods. The synthesis quality was evaluated using the peak signal-to-noise ratio (PSNR) and structural similarity index (SSIM) \cite{yi2019generative}. We also used Fréchet Inception Distance (FID) \cite{heusel2017gans} and Learned Perceptual Image Patch Similarity (LPIPS) \cite{zhang2018unreasonable} to evaluate image quality.

\noindent \textbf{Dataset.}
We used the ACRIN 6698 dataset \cite{newitt2019test, partridge2018diffusion}, which evaluates breast cancer response to NAC using DCE-MRI, focusing on the first phase after contrast injection. The dataset includes pre-treatment (Time 0), early-treatment (Time 1, 3 weeks after), and mid-treatment (Time 2, 12 weeks after) images. Out of 191 subjects, 148 were used for training and 43 for testing. Affine registration was applied to align Time 1 and Time 2 images to Time 0 before MIP using ANTs \cite{avants2009advanced}.

% We used the American College of Radiology Imaging Network (ACRIN) trial 6698 dataset for our study \cite{newitt2019test, partridge2018diffusion}. This dataset evaluates the effectiveness of DCE-MRI in assessing breast cancer response to NAC. The data had multiple sequential phases: one before contrast injection and five or more (for at least 8 minutes) after. We chose the first phase after contrast injection. We utilized DCE-MRI with a resolution of $256\times256\times80$ and a voxel size of $0.6055\times0.6055\times2.0 \text{mm}^3$. For training, $256\times256$ axial MIP of DCE-MRI was employed. The dataset includes pre-treatment (Time 0), early-treatment (Time 1, after 3 weeks of regimen 1 treatment), and mid-treatment (Time 2, after 12 weeks) images, as depicted in Fig \ref{fig1}. Clinical factors affecting the efficacy of NAC such as HR, HER2, and pCR, were used. HR and HER2 status are available before NAC, while pCR is confirmed after NAC. Still, we choose to include pCR because it is an established marker and there is an unmet need to simulate future imaging on a hypothetical pCR status. Out of 191 subjects, 148 were used for training and 43 for testing. Since each subject underwent MRI scans at multiple time points with varying geometry, we applied affine registration to align the Time 1 and Time 2 images to the Time 0 image before MIP using ANTs \cite{avants2009advanced}.

\begin{table} [t]
    \centering
    \caption{Comparison of our model's performance for Time 0 to Time 2 generation task. Bold denotes the best performance.}
    \label{table2}
    \vspace{3pt}
    \scalebox{0.65}{
    \begin{tabular}{ccccccccc}
        \toprule
        Task & \multicolumn{8}{c}{Time 0 $\rightarrow$ Time 2}\\
        \cmidrule(lr){2-9} 
        pCR & \multicolumn{4}{c}{pCR} & \multicolumn{4}{c}{Non-pCR} \\
        \cmidrule(lr){2-5} \cmidrule(lr){6-9}
        Model & PSNR $\uparrow$ & SSIM $\uparrow$ & FID $\downarrow$ & LPIPS $\downarrow$ & PSNR $\uparrow$ & SSIM $\uparrow$ & FID $\downarrow$ & LPIPS $\downarrow$ \\
        \midrule
        \multirow{1}{*}{Pix2pix}  \cite{isola2017image}       & 21.06 & 0.563 & 202.8 & 0.217 & 20.79 & 0.551 & 217.4 & 0.213 \\
        \multirow{1}{*}{CycleGAN} \cite{zhu2017unpaired}      & 21.27 & 0.568 & 215.9 & 0.233 & 20.99 & 0.561 & 195.9 & 0.233 \\
        \multirow{1}{*}{RegGAN}   \cite{NEURIPS2021_0f281810} & 20.65 & 0.529 & 222.4 & 0.252 & 20.53 & 0.522 & 195.7 & 0.239 \\
        \multirow{1}{*}{ResViT}   \cite{dalmaz2022resvit}     & 21.02 & 0.529 & 203.7 & 0.224 & 20.69 & 0.520 & 201.9 & 0.220 \\
        \multirow{1}{*}{LDM}      \cite{rombach2022high}      & 20.76 & 0.566 & 183.8 & 0.200 & 19.89 & 0.526 & 190.8 & 0.216 \\
        \multirow{1}{*}{\textbf{Ours}} & \textbf{21.65}     & \textbf{0.572} & \textbf{157.8} & \textbf{0.189} & \textbf{21.33} & \textbf{0.562} & \textbf{150.7} & \textbf{0.184} \\
        \bottomrule
    \end{tabular}
    }
    \vspace{-12pt}
\end{table}

\begin{figure*} [t]
    \vspace{-9pt}
    \centering
    \includegraphics[width=0.89\textwidth]{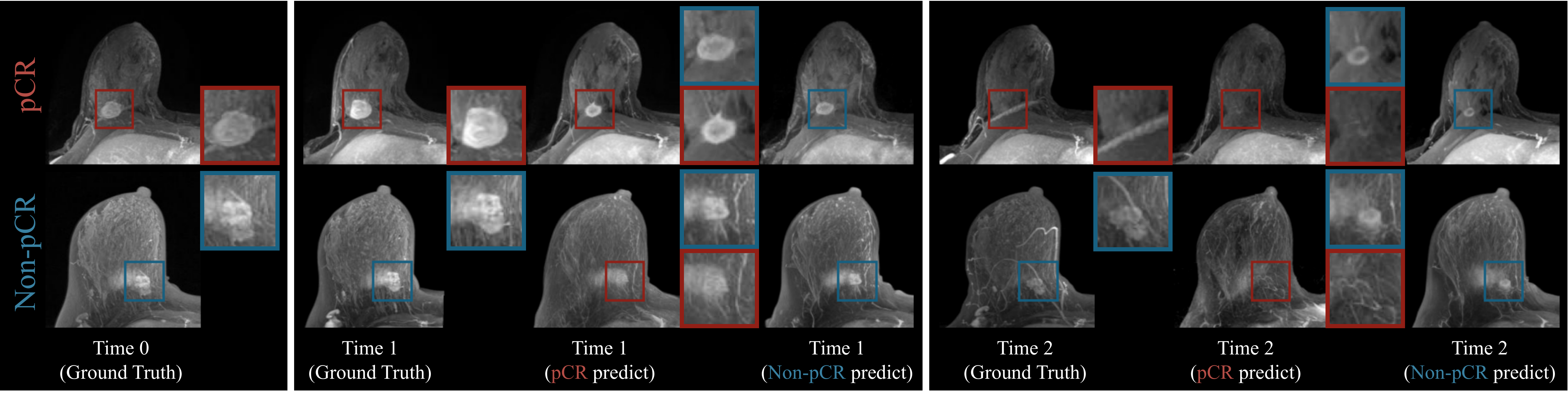}
    \vspace{-9pt}
    \caption{Illustration of image synthesis results using our model. Top: Simulation of Time 1 and Time 2 images for a pCR patient, showing results for both pCR and non-pCR prompts. Bottom: Simulation of Time 1 and Time 2 images for a non-pCR patient. The bounding boxes indicate tumor areas, with enlarged views shown above.} 
    \label{fig4}
    \vspace{-12pt}
\end{figure*}

\noindent \textbf{Comparisons of performance.}
We evaluated our model quantitatively using common metrics (PSNR, SSIM, FID \cite{heusel2017gans}, and LIPSIS \cite{zhang2018unreasonable}) and qualitatively comparing ours to other models trained to generate Time 1 or Time 2 images from Time 0. We used Pix2Pix \cite{isola2017image}, CycleGAN \cite{zhu2017unpaired}, RegGAN \cite{NEURIPS2021_0f281810}, ResViT \cite{dalmaz2022resvit}, and LDM \cite{rombach2022high} as comparison methods. Unlike others, our model handled both tasks with a single training. 

The qualitative evaluation for the Time 0 $\rightarrow$ Time 1 task (Fig \ref{fig3}) shows our model successfully reducing tumor size, particularly in pCR cases, while comparison models struggled. The quantitative results (Table \ref{table1}) show that alignment issues affected PSNR and SSIM for all models. Additionally, FID values may be inflated due to the limited dataset size. Despite this, our model outperformed others, demonstrating better simulation of post-NAC images compared to the comparison models. The qualitative evaluation of the Time 0 $\rightarrow$ Time 2 task, shown in Fig \ref{fig3}, highlights that while most GAN-based models failed to reduce tumor size, our model successfully depicted tumor shrinkage while preserving image quality. The quantitative results in Table \ref{table2} show our model achieving the highest performance across all metrics. Compared to Time 0 $\rightarrow$ Time 1, the clearer differences in Time 0 $\rightarrow$ Time 2 images made it easier for models to perform better, though our model remained the most accurate.

\setlength{\tabcolsep}{6pt}
\begin{table}[t]
    \centering
    \caption{Evaluation of tumor size changes reflected in comparison models.}
    \label{table3}
    \centering
    \vspace{3pt}
    \scalebox{0.65}{
    \begin{tabular}{ccccccccc}
        \toprule
         Task & \multicolumn{4}{c}{Time 0 $\rightarrow$ Time 1} & \multicolumn{4}{c}{Time 0 $\rightarrow$ Time 2}  \\
         \cmidrule(lr){2-5} \cmidrule(lr){6-9}
         pCR & \multicolumn{2}{c}{pCR} & \multicolumn{2}{c}{Non-pCR} & \multicolumn{2}{c}{pCR} & \multicolumn{2}{c}{Non-pCR} \\
        \cmidrule(lr){2-3} \cmidrule(lr){4-5} \cmidrule(lr){6-7} \cmidrule(lr){8-9}
        Model & Size & Match & Size  & Match & Size & Match & Size & Match  \\
        \midrule
        Pix2Pix  \cite{isola2017image}       & 70.7 & 52.3 & 61.3 & 57.8 & 78.8 & 90.3 & 77.6 & 69.0  \\
        CycleGAN \cite{zhu2017unpaired}      & 40.3 & 84.2 & 38.8 & 90.1 & 53.6 & 61.4 & 71.9 & 78.6  \\
        RegGAN   \cite{NEURIPS2021_0f281810} & 68.8 & 56.3 & 53.4 & 76.1 & 77.8 & 89.2 & 72.4 & 77.7 \\
        ResViT   \cite{dalmaz2022resvit}     & 57.3 & 80.2 & 49.0 & 86.1 & 62.8 & 71.9 & 67.1 & 86.7  \\
        LDM      \cite{rombach2022high}      & 51.4 & 92.7 & 57.1 & 67.4 & 66.3 & 76.0 & 74.6 & 74.0  \\
        \textbf{Ours} & 50.2 & \textbf{95.2} & 45.6 & \textbf{94.2} & 82.8 & \textbf{94.9} & 62.2 & \textbf{95.0}  \\
        \midrule
        Ground Truth & 47.93 & 100.0 & 43.13 & 100.0 & 87.48 & 100.0 & 59.28 & 100.0 \\
        \bottomrule
    \end{tabular}
    }
    \vspace{-9pt}
\end{table}

\noindent \textbf{Comparison of Changing Tumor Size.}
To evaluate if the generated images reflect tumor changes over time, we measured tumor size changes across models. Since the images lack tumor mask labels, we trained nnU-Net on the ground truth data. Metrics like DICE, which assess spatial overlap, are less suitable due to misalignment between time points. Instead, we focused on evaluating tumor size changes during NAC treatment. The trained model generated tumor masks, with the “size” metric representing tumor reduction relative to Time 0, and the “match” metric being the ratio of tumor size in the generated image to the ground truth. 

Table \ref{table3} shows tumor size changes for each model in the Time 0 $\rightarrow$ Time 1 task, divided by pCR and non-pCR. Tumor size decreased by 47.93\% in pCR cases and 43.13\% in non-pCR cases, with no significant difference due to the short time after NAC. Our model closely matched the ground truth, effectively capturing tumor changes (Fig \ref{fig4}).
For the Time 0 $\rightarrow$ Time 2 task, Table \ref{table3} shows significant differences between pCR and non-pCR cases, with pCR cases showing an 87.48\% tumor reduction and non-pCR cases a 59.28\% reduction. Our model had the highest match rate with ground truth. Fig \ref{fig4} confirms that the generated images accurately captured both conditions, with tumors almost disappearing in pCR cases and persisting but reducing in non-pCR cases.

\setlength{\tabcolsep}{6pt}
\begin{table}[t]
    \centering
    \caption{Ablation study for each component of the proposed model on the Time 0 $\rightarrow$ Time 1 task for pCR. 'Pretrained' indicates whether stable diffusion was pretrained, while 'Text' and 'Prompt Tuning' denote the type of prompt used.}
    \label{table4}
    \vspace{6pt}
    \scalebox{0.7}{
            \begin{tabular}{ccccccc}
                \toprule
                & \multicolumn{2}{c}{Prompt Type} & \multicolumn{4}{c}{Metrics} \\
                \cmidrule(lr){2-3} \cmidrule(lr){4-7}
                Pretrained & Text & Prompt tuning &  PSNR $\uparrow$ & SSIM $\uparrow$ & FID $\downarrow$ & LPIPS $\downarrow$  \\
                \midrule
                           & \checkmark & & 18.094 & 0.5196 & 215.45 & 0.2127 \\
                \checkmark &            & & 18.214 & 0.5241 & 220.28 & 0.2145 \\
                \checkmark & \checkmark & & 19.241 & 0.5463 & 197.53 & 0.1948 \\
                \checkmark & & \checkmark & \textbf{20.279} & \textbf{0.5619} & \textbf{175.45} & \textbf{0.1868} \\
                \bottomrule
            \end{tabular}
    }
    \vspace{-6pt}
\end{table}

\noindent \textbf{Ablation Study.}
We conducted an ablation study for the Time 0 $\rightarrow$ Time 1 task in Table \ref{table4}. Without the pretrained stable diffusion model, performance was similar to comparison models. Although omitting the prompt improved PSNR and SSIM, it decreased FID and LPIPS. This shows that using prompts with clinical data like pCR, HR, and HER2 is crucial. Our method with prompt tuning achieved the best performance, demonstrating its effectiveness in simulating NAC responses and integrating medical context.

% We conducted an ablation study to assess the impact of each component of our method on performance for the Time 0 $\rightarrow$ Time 1 task in Table \ref{table4}. Using the original text prompt without the pretrained stable diffusion model, the performance was similar to that of the comparison models in Table \ref{table1}. Using the pretrained stable diffusion model with partial fine-tuning and training with the text prompt improved performance slightly. Even though we used a pretrained model, omitting the prompt led to slightly better performance in PSNR and SSIM compared to models that did not use the pretrained model. However, performance decreased in FID and LPIPS. This indicates that incorporating clinical information such as pCR, HR, and HER2 through prompts is crucial for achieving improved performance. Our method, which employed prompt tuning, achieved the highest quantitative performance. This demonstrates that our proposed method effectively simulates NAC responses and that prompt tuning successfully incorporates medical context not initially learned by the text encoder into the generative model.

\section{Discussion}

\noindent \textbf{Limitations.} First is the small sample size, making accurate simulation difficult, a common issue in medical imaging. Another is the variation in environments for breast MRI scans at different time points, making alignment of complex structures like blood vessels difficult. We believe improved deep learning-based alignment techniques could address this. Despite imperfect alignment in this study, we demonstrated the potential of our method by successfully simulating post-NAC images and observing tumor changes.

\noindent \textbf{Conclusion.}
We used a diffusion model to simulate post-NAC images from pre-treatment data, incorporating patient-specific clinical factors like pCR, HR, and HER2 through prompt tuning. This allowed fine-tuning of generative models that lack medical context. Our method outperformed, as confirmed by the ablation study. We expect this approach can be integrated into clinical workflows to help predict patient prognosis and improve personalized treatment strategies.

\clearpage

\section{Acknowledgments}
% \noindent \textbf{Acknowledgments.}
This study was supported by National Research Foundation (RS-2024-00408040), Institute for Basic Science (IBS-R015-D2), AI Graduate School Support Program (Sungkyunkwan University) (RS-2019-II190421), ICT Creative Consilience program (IITP-2025-RS-2020-II201821), and the Artificial Intelligence Innovation Hub program (RS-2021-II212068).

\bibliographystyle{IEEEbib}
\bibliography{strings,refs}

\begin{thebibliography}{10}

\bibitem{lei2021global}
Shaoyuan Lei et~al.,
\newblock ``Global patterns of breast cancer incidence and mortality: A population-based cancer registry data analysis from 2000 to 2020,''
\newblock {\em Cancer Communications}, vol. 41, no. 11, pp. 1183--1194, 2021.

\bibitem{mieog2007neoadjuvant}
JSD Mieog et~al.,
\newblock ``Neoadjuvant chemotherapy for operable breast cancer,''
\newblock {\em Journal of British Surgery}, vol. 94, 2007.

\bibitem{spring2020pathologic}
Laura~M Spring et~al.,
\newblock ``Pathologic complete response after neoadjuvant chemotherapy and impact on breast cancer recurrence and survival: a comprehensive meta-analysis,''
\newblock {\em Clinical cancer research}, vol. 26, no. 12, pp. 2838--2848, 2020.

\bibitem{huang2020association}
Min Huang et~al.,
\newblock ``Association of pathologic complete response with long-term survival outcomes in triple-negative breast cancer: a meta-analysis,''
\newblock {\em Cancer Research}, vol. 80, no. 24, pp. 5427--5434, 2020.

\bibitem{kim2023multi}
Jonghun Kim and Hyunjin Park,
\newblock ``Multi-modal cross attention network for predicting pathological complete response in breast cancer mri,''
\newblock in {\em 2023 5th International Conference on Control and Robotics (ICCR)}. IEEE, 2023, pp. 250--254.

\bibitem{10635671}
Jonghun Kim and Hyunjin Park,
\newblock ``Radiomics-guided multimodal self-attention network for predicting pathological complete response in breast mri,''
\newblock in {\em 2024 IEEE International Symposium on Biomedical Imaging (ISBI)}, 2024, pp. 1--5.

\bibitem{frenel2023efficacy}
J-S Frenel et~al.,
\newblock ``Efficacy of front-line treatment for hormone receptor-positive her2-negative metastatic breast cancer with germline brca1/2 mutation,''
\newblock {\em British Journal of Cancer}, vol. 128, no. 11, pp. 2072--2080, 2023.

\bibitem{yankeelov2007dynamic}
Thomas~E Yankeelov and John~C Gore,
\newblock ``Dynamic contrast enhanced magnetic resonance imaging in oncology: theory, data acquisition, analysis, and examples,''
\newblock {\em Current Medical Imaging}, vol. 3, no. 2, pp. 91--107, 2007.

\bibitem{kuhl2014abbreviated}
Christiane~K Kuhl et~al.,
\newblock ``Abbreviated breast magnetic resonance imaging (mri): first postcontrast subtracted images and maximum-intensity projection—a novel approach to breast cancer screening with mri,''
\newblock {\em Journal of Clinical Oncology}, vol. 32, no. 22, pp. 2304--2310, 2014.

\bibitem{ho2020denoising}
Jonathan Ho et~al.,
\newblock ``Denoising diffusion probabilistic models,''
\newblock {\em Advances in neural information processing systems}, 2020.

\bibitem{goodfellow2014generative}
Ian Goodfellow et~al.,
\newblock ``Generative adversarial nets,''
\newblock {\em Advances in neural information processing systems}, vol. 27, 2014.

\bibitem{dhariwal2021diffusion}
Prafulla Dhariwal and Alexander Nichol,
\newblock ``Diffusion models beat gans on image synthesis,''
\newblock {\em Advances in neural information processing systems}, vol. 34, pp. 8780--8794, 2021.

\bibitem{Kim_2024_WACV}
Jonghun Kim and Hyunjin Park,
\newblock ``Adaptive latent diffusion model for 3d medical image to image translation: Multi-modal magnetic resonance imaging study,''
\newblock in {\em Proceedings of the IEEE/CVF Winter Conference on Applications of Computer Vision (WACV)}, January 2024, pp. 7604--7613.

\bibitem{na2024radiomicsfill}
Inye Na et~al.,
\newblock ``Radiomicsfill-mammo: Synthetic mammogram mass manipulation with radiomics features,''
\newblock {\em arXiv preprint arXiv:2407.05683}, 2024.

\bibitem{isensee2021nnu}
Fabian Isensee et~al.,
\newblock ``nnu-net: a self-configuring method for deep learning-based biomedical image segmentation,''
\newblock {\em Nature methods}, vol. 18, no. 2, pp. 203--211, 2021.

\bibitem{zhang2023adding}
Lvmin Zhang et~al.,
\newblock ``Adding conditional control to text-to-image diffusion models,''
\newblock in {\em Proceedings of the IEEE/CVF International Conference on Computer Vision}, 2023.

\bibitem{radford2021learning}
Alec Radford et~al.,
\newblock ``Learning transferable visual models from natural language supervision,''
\newblock in {\em International conference on machine learning}. PMLR, 2021, pp. 8748--8763.

\bibitem{jia2022visual}
Menglin Jia et~al.,
\newblock ``Visual prompt tuning,''
\newblock in {\em European Conference on Computer Vision}. Springer, 2022, pp. 709--727.

\bibitem{isola2017image}
Phillip Isola et~al.,
\newblock ``Image-to-image translation with conditional adversarial networks,''
\newblock in {\em Proceedings of the IEEE conference on computer vision and pattern recognition}, 2017.

\bibitem{zhu2017unpaired}
Jun-Yan Zhu et~al.,
\newblock ``Unpaired image-to-image translation using cycle-consistent adversarial networks,''
\newblock in {\em Proceedings of the IEEE international conference on computer vision}, 2017.

\bibitem{NEURIPS2021_0f281810}
Lingke Kong et~al.,
\newblock ``Breaking the dilemma of medical image-to-image translation,''
\newblock in {\em Advances in Neural Information Processing Systems}. 2021, vol.~34, pp. 1964--1978, Curran Associates, Inc.

\bibitem{dalmaz2022resvit}
Onat Dalmaz et~al.,
\newblock ``Resvit: residual vision transformers for multimodal medical image synthesis,''
\newblock {\em IEEE Transactions on Medical Imaging}, vol. 41, no. 10, pp. 2598--2614, 2022.

\bibitem{rombach2022high}
Robin Rombach et~al.,
\newblock ``High-resolution image synthesis with latent diffusion models,''
\newblock in {\em Proceedings of the IEEE/CVF conference on computer vision and pattern recognition}, 2022.

\bibitem{loshchilov2017decoupled}
Ilya Loshchilov et~al.,
\newblock ``Decoupled weight decay regularization,''
\newblock {\em arXiv preprint arXiv:1711.05101}, 2017.

\bibitem{song2020denoising}
Jiaming Song et~al.,
\newblock ``Denoising diffusion implicit models,''
\newblock {\em arXiv preprint arXiv:2010.02502}, 2020.

\bibitem{newitt2019test}
David Newitt et~al.,
\newblock ``Test--retest repeatability and reproducibility of adc measures by breast dwi: Results from the acrin 6698 trial,''
\newblock {\em Journal of Magnetic Resonance Imaging}, vol. 49, no. 6, pp. 1617--1628, 2019.

\bibitem{partridge2018diffusion}
Partridge et~al.,
\newblock ``Diffusion-weighted mri findings predict pathologic response in neoadjuvant treatment of breast cancer: the acrin 6698 multicenter trial,''
\newblock {\em Radiology}, vol. 289, no. 3, pp. 618--627, 2018.

\bibitem{avants2009advanced}
Avants et~al.,
\newblock ``Advanced normalization tools (ants),''
\newblock {\em Insight j}, vol. 2, no. 365, pp. 1--35, 2009.

\bibitem{heusel2017gans}
Martin Heusel et~al.,
\newblock ``Gans trained by a two time-scale update rule converge to a local nash equilibrium,''
\newblock {\em Advances in neural information processing systems}, vol. 30, 2017.

\bibitem{zhang2018unreasonable}
Richard Zhang et~al.,
\newblock ``The unreasonable effectiveness of deep features as a perceptual metric,''
\newblock in {\em Proceedings of the IEEE conference on computer vision and pattern recognition}, 2018.

\end{thebibliography}

\end{document}